\documentclass[preprint,12pt]{elsarticle}

\usepackage{amssymb}

\usepackage{lineno}  
\usepackage{amsmath} 
\usepackage{subfigure}
\usepackage{multirow} 
\usepackage{booktabs} 
\usepackage[normalem]{ulem} 
\usepackage[dvipsnames]{xcolor}
\definecolor{mygray}{gray}{0.6}
\usepackage{siunitx} 

\journal{}

\begin{document}
\begin{frontmatter}



\title{ClusterDDPM: An EM clustering framework with Denoising Diffusion Probabilistic Models}

\author{Jie Yan}
\author{Jing Liu}
\author{Zhong-yuan Zhang\corref{mycorrespondingauthor}}
\address{School of Statistics and Mathematics, \\ Central University of Finance and Economics, Beijing, P.R.China}
\cortext[mycorrespondingauthor]{Corresponding author}
\ead{zhyuanzh@gmail.com}

%

\begin{abstract}
Variational autoencoder (VAE) and generative adversarial networks (GAN) have found widespread applications in clustering and have achieved significant success. However, the potential of these approaches may be limited due to VAE's mediocre generation capability or GAN's well-known instability during adversarial training. In contrast, denoising diffusion probabilistic models (DDPMs) represent a new and promising class of generative models that may unlock fresh dimensions in clustering. In this study, we introduce an innovative expectation-maximization (EM) framework for clustering using DDPMs. In the E-step, we aim to derive a mixture of Gaussian priors for the subsequent M-step. In the M-step, our focus lies in learning clustering-friendly latent representations for the data by employing the conditional DDPM and matching the distribution of latent representations to the mixture of Gaussian priors. We present a rigorous theoretical analysis of the optimization process in the M-step, proving that the optimizations are equivalent to maximizing the lower bound of the Q function within the vanilla EM framework under certain constraints. Comprehensive experiments validate the advantages of the proposed framework, showcasing superior performance in clustering, unsupervised conditional generation and latent representation learning.
\end{abstract}


\begin{highlights}
\item
We propose an innovative EM clustering framework with denoising diffusion probabilistic models (DDPMs). As far as we can ascertain, this is the first exploration of DDPMs in clustering.

\item
We provide a rigorous theoretical analysis of the optimization process within the proposed framework's M-step.

\item
Comprehensive experiments validate the significant advantages of our proposed ClusterDDPM framework and offer valuable insights into its performance.
\end{highlights}


\end{frontmatter}


\section{Introduction}
Clustering has witnessed remarkable progress in handling complex high-dimensional data, particularly images, owing to the powerful representation learning capabilities of deep generative models. Notably, recent advancements \cite{vade, mukherjee2019clustergan, gan-em, de2022top, yang2020clustering} have extensively employed two commonly used models: the variational autoencoder (VAE) \cite{kingma2013auto} and the generative adversarial network (GAN) \cite{kingma2013auto}. However, it is widely acknowledged that VAE's generative capacity is modest \cite{pandey2021vaes}, and GAN's adversarial training is notably unstable \cite{metz2016unrolled}, thereby limiting their potential.

In contrast, denoising diffusion probabilistic models (DDPMs) \cite{sohl2015deep, ho2020denoising} have emerged as a new class of generative models, surpassing GANs in various generation tasks such as image generation \cite{hu2022global}, video generation \cite{ni2023conditional}, and graph generation \cite{ye2022first}. Beyond their impressive generative capabilities, DDPMs have demonstrated remarkable representation learning abilities \cite{preechakul2022diffusion, zhang2022unsupervised}, which may open new opportunities of deep generative models in clustering. However, these opportunities have remained unexplored in existing literatures.

\begin{figure}[!t]
\centering
\includegraphics[height = 8cm, width = 10cm]{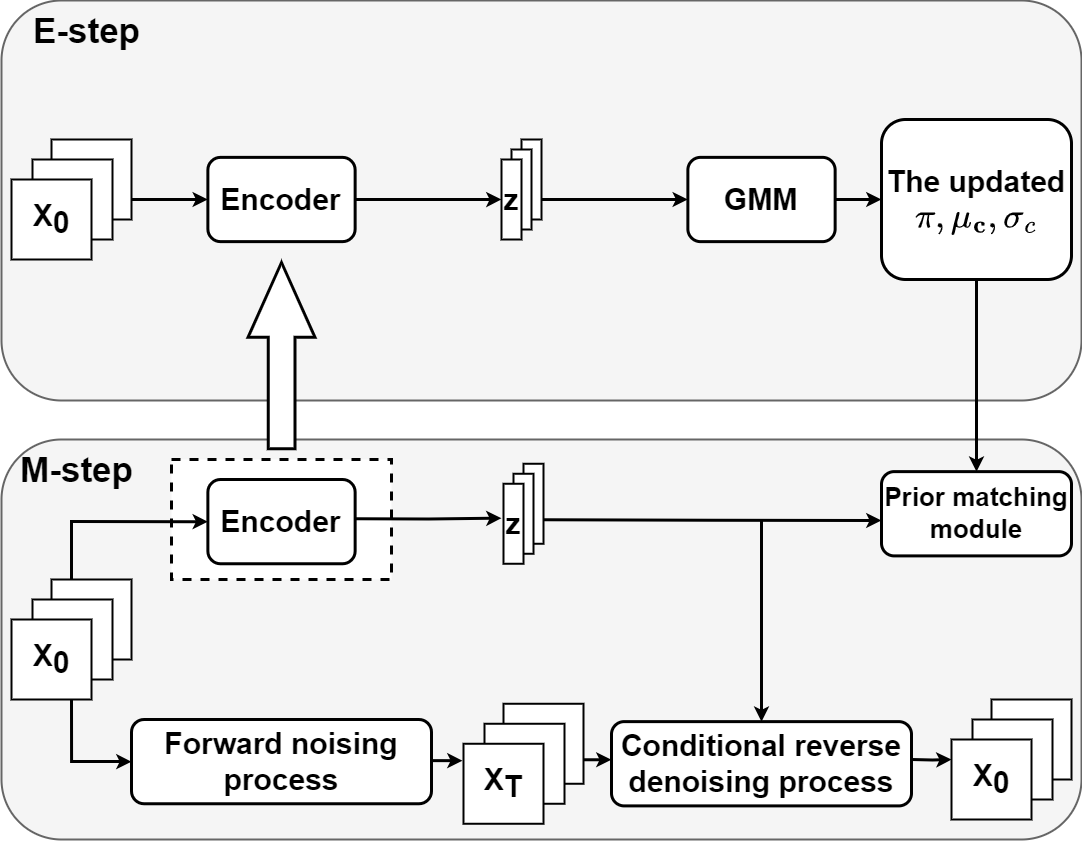}
\caption{ClusterDDPM architecture. $\mathbf{X}_0$: clear images, $\mathbf{X}_T$: noising images, $z$: the latent representation of $\mathbf{X}_0$, $\left\{\boldsymbol{\pi},\, \boldsymbol{\mu}_c,\, \boldsymbol{\sigma}_c\right\}$ are the parameters of gaussian mixture model (GMM) and $c$ is the cluster indicator. The conditional DDPM aims to encourage the encoder to encode the information of $\mathbf{X}_0$ in $z$, while the prior matching module aims to regularize $z$ to reside on the mixture of Gaussians manifold. The final clustering result can be obtained by performing GMM on the learned latent representations.}
\label{framework}
\end{figure}

To address this gap, we propose \textbf{ClusterDDPM}, an innovative expectation-maximization (EM) clustering framework with DDPMs. As shown in Fig. \ref{framework}, the E-step of our framework aims to derive a mixture of Gaussian priors for the M-step by performing Gaussian mixture model (GMM) \cite{mclachlan1988mixture} in the latent space. Subsequently, the M-step endeavors to learn GMM-friendly latent representations for the data by employing the conditional DDPM and matching the distribution of latent representations with the mixture of Gaussian priors. The final clustering result can be obtained by performing GMM on the learned latent representations. We provide a rigorous theoretical analysis of the optimization process in the M-step, proving that the optimizations are equivalent to maximizing the lower bound of the Q function \cite{bishop2006pattern} under certain constraints. Furthermore, we offer insights into the effectiveness of the learned latent representations, supported by qualitative and quantitative analyses of ClusterDDPM's performance. Comprehensive experiments validate the advantages of the proposed model in terms of clustering, unsupervised conditional generation and latent representation learning.

In summary, this work makes the following contributions:
\begin{itemize}
\item
We propose an innovative EM clustering framework with DDPMs. As far as we can ascertain, this is the first exploration of DDPMs in clustering.

\item
We provide a rigorous theoretical analysis of the optimization process within the proposed framework's M-step.

\item
Comprehensive experiments validate the significant advantages of our proposed ClusterDDPM framework and offer valuable insights into its performance.

\end{itemize}

\section{Related work}
Generative models, particularly variational autoencoder (VAE) \cite{kingma2013auto} and generative adversarial networks (GAN) \cite{gan}, have showcased substantial progress in improving clustering tasks by acquiring more clustering-friendly representations. This progress has led to the emergence of two distinct categories of representative approaches for clustering applications. The first type utilizes the generative model to create pseudo-labeled generated samples, which are then used to train a classifier. This trained classifier is subsequently employed for clustering real samples. Notable examples include GAN mixture model (GANMM) \cite{yu2018mixture}, GAN-based EM learning framework (GAN-EM) \cite{gan-em} ClusterGAN \cite{mukherjee2019clustergan} and hierarchical clustering with multi-generator GANs (HC-MGAN) \cite{de2022top}. The second type combines the generative model with the Gaussian mixture model (GMM) to learn GMM-friendly latent representations, enabling the derivation of $K$ Gaussian components for clustering assignments based on these representations. Prominent examples of this approach include variational deep embedding (VaDE) \cite{vade} and VaGAN-GMM \cite{yang2020clustering}, which are the most related works to ours. While these methods have shown successful applications, their potential might be hampered by the mediocre generation capability of VAE or the well-known instability of GAN during adversarial training.

Recently, a new class of generative models, denoising diffusion probabilistic models (DDPMs) \cite{sohl2015deep, ho2020denoising}, has emerged and shattered the monopoly of GANs in high-fidelity sample generation. The basic idea of them is to match a data distribution $q(\mathbf{x}_0)$ by learning to reverse a pre-defined, gradual, multi-step forward noising process. DDPMs can be regarded as a special type of VAE, where their forward noising process corresponds to the encoding process in VAE, and the reverse denoising process corresponds to the decoding process \cite{croitoru2023diffusion}. Formally, the forward noising process gradually disrupts $q(\mathbf{x}_0)$ into $\mathcal{N}\left(\mathbf{0}, \, \mathbf{I}\right)$ by adding Gaussian noises with
a fixed variance schedule $\{\beta\}_{t = 1}^T$, which is defined as:
\begin{align}
q(\mathbf{x}_{1:T} | \mathbf{x}_0)  & = \prod_{t=1}^T q\left(\mathbf{x}_{t} | \mathbf{x}_{t - 1}\right), \label{q_infer}\\
q\left(\mathbf{x}_{t} | \mathbf{x}_{t - 1}\right) & = \mathcal{N}\left(\sqrt{1-\beta_t} \mathbf{x}_{t - 1}, \,\beta_t \mathbf{I}\right).
\end{align}
By recursively applying the reparameterization trick to $q\left(\mathbf{x}_{t} | \mathbf{x}_{t - 1}\right)$, one can obtain:
\begin{equation}
q\left(\mathbf{x}_{t} | \mathbf{x}_0\right) = \mathcal{N}\left(\sqrt{\bar{\alpha}_t} \mathbf{x}_0, \, (1-\bar{\alpha}_t) \mathbf{I}\right),
\end{equation}
which enables us to sample $\mathbf{x}_t$ from $\mathbf{x}_0$  for any timestep $t$, $\bar{\alpha}_t = \prod_{s=1}^t \alpha_s$ and $\alpha_s = 1- \beta_s$. Then, one can use the Markov property to obtain $q\left(\mathbf{x}_{t} | \mathbf{x}_{t - 1}, \mathbf{x}_0\right) = q\left(\mathbf{x}_{t} | \mathbf{x}_{t - 1}\right)$, and use the Bayes theorem to obtain the true posterior:
\begin{equation}
q\left(\mathbf{x}_{t-1} | \mathbf{x}_{t}, \mathbf{x}_0\right) = \mathcal{N}\left(\mathbf{x}_{t-1} | \boldsymbol{\mu}_q\left(\mathbf{x}_t, \mathbf{x}_0\right), \boldsymbol{\Sigma}_q\left(t\right)\right),
\label{true_posterior}
\end{equation}
where $\boldsymbol{\mu}_q\left(\mathbf{x}_t, \mathbf{x}_0\right) = \frac{1}{\sqrt{\alpha_t}}\left(\mathbf{x}_t-\frac{\beta_t}{\sqrt{1-\bar{\alpha}_t}} \boldsymbol{\epsilon}\right)$, $\epsilon$ is the noise added to $\mathbf{x}_0$ to obtain $\mathbf{x}_t$ and $\epsilon \sim \mathcal{N}\left(\mathbf{0}, \, \mathbf{I}\right)$, and $\boldsymbol{\Sigma}_q\left(t\right) = \frac{1-\bar{\alpha}_{t-1}}{1-\bar{\alpha}_t} \beta_t$. For the reverse denoising process, it is parameterized to match the true posterior, and is modeled as:
\begin{equation}
p_{\boldsymbol{\theta}}\left(\mathbf{x}_{t-1} | \mathbf{x}_t\right) = \mathcal{N}\left(\mathbf{x}_{t-1} | \boldsymbol{\mu}_{\boldsymbol{\theta}}\left(\mathbf{x}_t, t\right), \boldsymbol{\Sigma}_{\boldsymbol{\theta}}\left(t\right)\right).
\label{p_theta}
\end{equation}
In the DDPM \cite{ho2020denoising}, $\boldsymbol{\mu}_{\boldsymbol{\theta}}\left(\mathbf{x}_t, t\right)$ is parameterized by a noise prediction network $\boldsymbol{\epsilon}_{\boldsymbol{\theta}}\left(\mathbf{x}_t, t\right)$:
\begin{equation}
\boldsymbol{\mu}_{\boldsymbol{\theta}}\left(\mathbf{x}_t, t\right) = \frac{1}{\sqrt{\alpha_t}}\left(\mathbf{x}_t-\frac{\beta_t}{\sqrt{1-\bar{\alpha}_t}} \boldsymbol{\epsilon}_{\boldsymbol{\theta}}\left(\mathbf{x}_t, t\right)\right),
\label{mu_no_cond}
\end{equation}
and $\boldsymbol{\Sigma}_{\boldsymbol{\theta}}\left(t\right)$ is an untrained, time-dependent constant:
\begin{equation}
\boldsymbol{\Sigma}_{\boldsymbol{\theta}}\left(t\right) = \beta_t\mathbf{I}.
\label{sigma_no_cond}
\end{equation}
Finally, this model can be trained by minimizing the following simplified training objective:
\begin{equation}
\mathcal{L}_{simple} = E_{t, \mathbf{x}_0, \boldsymbol{\epsilon}}\left[\left\|\boldsymbol{\epsilon} - \boldsymbol{\epsilon}_{\boldsymbol{\theta}}\left(\mathbf{x}_t, t\right)\right\|^2\right],
\end{equation}
where $\mathbf{x}_t = \sqrt{\bar{\alpha}_t} \mathbf{x}_0+\sqrt{1-\bar{\alpha}_t} \boldsymbol{\epsilon}$.

Beyond their impressive generative capabilities, DDPMs have demonstrated remarkable representation learning abilities \cite{preechakul2022diffusion, zhang2022unsupervised}, which may potentially unlock new opportunities for deep generative models in clustering. However, the exploration of these opportunities remains uncharted. In this work, we propose an innovative expectation-maximization (EM) clustering framework with DDPMs. Comprehensive experiments validate the advantages of it in terms of clustering, unsupervised conditional generation and latent representation learning.

\section{ClusterDDPM}
In this section, we first detail the generative process of ClusterDDPM and then derive its objective function based on this process. Subsequently, we introduce an EM learning algorithm to solve the optimization problem and present the theoretical analysis of it.


\subsection{The generative process}
In ClusterDDPM, we follow the assumption of GMM that the data points are drawn from a distribution of Gaussian mixtures, and we generate samples in a conditional denoising process \cite{ho2022classifier, preechakul2022diffusion} fashion. The generative process is modeled as follows:

\begin{enumerate}
\item
To generate a sample $\mathbf{x}_0$ in cluster $c$, we first select a cluster indicator $c$, following a categorical distribution parameterized by $\boldsymbol{\pi}$:
\begin{equation}
p(c) =  \operatorname{Cat}(c | \boldsymbol{\pi}),
\label{p_c}
\end{equation}
where $\boldsymbol{\pi} =\left(\pi_c\right)_{c = 1}^K$, $\pi_c$ is the prior probability for cluster $c$ and $K$ is the number of clusters.

\item
Then, we select a low-dimensional vector $\mathbf{z}$ from cluster $c$ to control the generated content of $\mathbf{x}_0$. The probability distribution of $\mathbf{z}$ is defined as:
\begin{equation}
p(\mathbf{z} | c) =  \mathcal{N}\left(\mathbf{z} | \boldsymbol{\mu}_c,\, \operatorname{diag}(\boldsymbol{\sigma}_c^2)\right),
\label{p_z_c}
\end{equation}
where $\boldsymbol{\mu}_c$ and $\boldsymbol{\sigma}_c^2$ are the mean and variance of the $c$-th Gaussian component, respectively.

\item
Finally, we initiate a pure Gaussian noise $\mathbf{x_T}$, progressively denoise it based on the low-dimensional vector $\mathbf{z}$ until it becomes the completely denoised sample $\mathbf{x_0}$.
\begin{itemize}
\item
Select a pure Gaussian noise $\mathbf{x_T}$ from the probability distribution:
\begin{equation}
p(\mathbf{x}_T) = \mathcal{N}\left(\mathbf{x}_T | \mathbf{0}, \, \mathbf{I}\right)
\end{equation}
\item
Conditional reverse denoising process. For each conditional denoising transition step, we model $p_{\boldsymbol{\theta}}\left(\mathbf{x}_{t-1} | \mathbf{x}_t, \mathbf{z}\right)$ as an approximation to the ground-truth denoising transition $q\left(\mathbf{x}_{t-1} | \mathbf{x}_t, \mathbf{x}_0 \right)$ defined in Equation (\ref{true_posterior}). When $t > 1$, each conditional denoising step is defined as:
\begin{equation}
p_{\boldsymbol{\theta}}\left(\mathbf{x}_{t-1} | \mathbf{x}_t, \mathbf{z}\right) = \mathcal{N}\left(\mathbf{x}_{t-1} | \boldsymbol{\mu}_{\boldsymbol{\theta}}\left(\mathbf{x}_t, t, \mathbf{z}\right), \beta_t\mathbf{I}\right),
\end{equation}
where $\boldsymbol{\mu}_{\boldsymbol{\theta}}\left(\mathbf{x}_t, t, \mathbf{z}\right)$ is parameterized by a noise prediction network $\boldsymbol{\epsilon}_{\boldsymbol{\theta}}\left(\mathbf{x}_t, t, \mathbf{z}\right)$:
\begin{equation}
\boldsymbol{\mu}_{\boldsymbol{\theta}}\left(\mathbf{x}_t, t, \mathbf{z}\right) = \frac{1}{\sqrt{\alpha_t}}\left(\mathbf{x}_t-\frac{\beta_t}{\sqrt{1-\bar{\alpha}_t}} \boldsymbol{\epsilon}_{\boldsymbol{\theta}}\left(\mathbf{x}_t, t, \mathbf{z}\right)\right).
\end{equation}
When $t = 1$, the conditional denoising step is defined as:
\begin{equation}
p_{\boldsymbol{\theta}}\left(\mathbf{x}_{0} | \mathbf{x}_1, \mathbf{z}\right) = \mathcal{N}\left(\mathbf{x}_{0} | \boldsymbol{\mu}_{\boldsymbol{\theta}}\left(\mathbf{x}_1, 1, \mathbf{z}\right), \mathbf{0}\right).
\end{equation}
\end{itemize}
\end{enumerate}
According to the above generative process, our generative model is presented as:
\begin{equation}
p\left(\mathbf{x}_{0:T}, \mathbf{z}, c\right) = \prod_{t=1}^T p_{\boldsymbol{\theta}}\left(\mathbf{x}_{t-1} | \mathbf{x}_t, \mathbf{z}\right)p\left(\mathbf{x}_T\right)p(\mathbf{z} | c)p(c),
\label{p_joint}
\end{equation}
where $\mathbf{x}_{t}\,(t < T)$ and $c$ are conditionally independent given $\mathbf{z}$.

\subsection{The objective function}
A ClusterDDPM instance is trained to maximize the likelihood of the provided data points. For the generative model defined in Equation (\ref{p_joint}),  the evidence lower bound (ELBO) of the log-likelihood can be derived using Jensen’s inequality as follows:
\begin{align}
\log p(\mathbf{x}_{0}) & =\log \int_{\mathbf{x}_{1:T}} \int_{\mathbf{z}} \sum_c p\left(\mathbf{x}_{0:T}, \mathbf{z}, c\right) d \mathbf{z} d \mathbf{x}_{1:T} \nonumber \\
& \geq E_{q_{\boldsymbol{\phi}}(\mathbf{x}_{1:T}, \mathbf{z}, c | \mathbf{x}_0)}\left[\log \frac{p\left(\mathbf{x}_{0:T}, \mathbf{z}, c\right)}{q_{\boldsymbol{\phi}}(\mathbf{x}_{1:T}, \mathbf{z}, c | \mathbf{x}_0)}\right] \nonumber \\
& = \mathcal{L}_{\mathrm{ELBO}}(\mathbf{x}_0),
\label{elbo}
\end{align}
where $q_{\boldsymbol{\phi}}(\mathbf{x}_{1:T}, \mathbf{z}, c | \mathbf{x}_0)$ is the approximation of the ground-truth posterior $p(\mathbf{x}_{1:T}, \mathbf{z}, c | \mathbf{x}_0)$. By the mean-field distribution assumption, the approximated posterior can be further factorized as:
\begin{equation}
q_{\boldsymbol{\phi}}(\mathbf{x}_{1:T}, \mathbf{z}, c | \mathbf{x}_0) = q(\mathbf{x}_{1:T} | \mathbf{x}_0)q_{\boldsymbol{\phi}}(\mathbf{z} | \mathbf{x}_0)q(c | \mathbf{x}_0).
\label{q_joint}
\end{equation}
In ClusterDDPM, the forward noising process $q(\mathbf{x}_{1:T} | \mathbf{x}_0)$ (already defined in Equation (\ref{q_infer})) is a pre-defined linear Gaussian model, which is not learned. For $q_{\boldsymbol{\phi}}(\mathbf{z} | \mathbf{x}_0)$, we define it as:
\begin{equation}
q_{\boldsymbol{\phi}}(\mathbf{z} | \mathbf{x}_0) = \mathcal{N}\left(\mathbf{z} | \boldsymbol{\mu_{\phi}},\, \operatorname{diag}(\boldsymbol{\sigma}_{\boldsymbol{\phi}}^2)\right),
\end{equation}
and use a encoder network $\boldsymbol{f_{\phi}}$ to parameterize $\boldsymbol{\mu_{\phi}}$ and ${\boldsymbol{\sigma}_{\boldsymbol{\phi}}^2}$:
\begin{equation}
\left[\boldsymbol{\mu_{\phi}}; \log {\boldsymbol{\sigma}_{\boldsymbol{\phi}}^2}\right] = \boldsymbol{f_{\phi}}(\mathbf{x}_0).
\label{q_mu_sigma}
\end{equation}
By applying the reparameterization trick, the latent representation $\mathbf{z}$ can be obtained by:
\begin{equation}
\mathbf{z} = \boldsymbol{\mu_{\phi}} + \boldsymbol{\sigma}_{\boldsymbol{\phi}} \circ \boldsymbol{\epsilon},
\label{q_z}
\end{equation}
where $\boldsymbol{\epsilon} \sim \mathcal{N}\left(\mathbf{0}, \, \mathbf{I}\right)$ and $\circ$ is element-wise multiplication.
For $q(c | \mathbf{x}_0)$, it has a closed-form solution, derived by substituting Equations (\ref{p_joint}) and (\ref{q_joint}) into Equation (\ref{elbo}):
\begin{alignat}{2}
\mathcal{L}_{\mathrm{ELBO}}(\mathbf{x}_0) &= &&\,E_{q_{\boldsymbol{\phi}}(\mathbf{x}_{1:T}, \mathbf{z}, c | \mathbf{x}_0)}\left[\log \frac{p\left(\mathbf{x}_{0:T} | \mathbf{z}\right)p(\mathbf{z})}{q(\mathbf{x}_{1:T} | \mathbf{x}_0)q_{\boldsymbol{\phi}}(\mathbf{z} | \mathbf{x}_0)}\right] \nonumber \\[2mm]
& &&+ E_{q_{\boldsymbol{\phi}}(\mathbf{x}_{1:T}, \mathbf{z}, c | \mathbf{x}_0)}\left[\log \frac{p(c|\mathbf{z})}{q(c | \mathbf{x}_0)}\right] \nonumber \\[2mm]
&=  &&\,E_{q_{\boldsymbol{\phi}}(\mathbf{x}_{1:T}, \mathbf{z} | \mathbf{x}_0)}\left[\log \frac{p\left(\mathbf{x}_{0:T} | \mathbf{z}\right)p(\mathbf{z})}{q(\mathbf{x}_{1:T} | \mathbf{x}_0)q_{\boldsymbol{\phi}}(\mathbf{z} | \mathbf{x}_0)}\right] \nonumber \\[2mm] & &&- E_{q_{\boldsymbol{\phi}}(\mathbf{z} | \mathbf{x}_0)}\left[D_{KL}\left(q(c | \mathbf{x}_0) || p(c|\mathbf{z})\right)\right],
\label{q(c|x)}
\end{alignat}
where $D_{KL}(\cdot || \cdot)$ is the Kullback–Leibler divergence. To maximize $\mathcal{L}_{\mathrm{ELBO}}(\mathbf{x}_0)$, $D_{KL}\left(q(c | \mathbf{x}_0) || p(c|\mathbf{z})\right) \equiv 0$ should hold true, as the first term in Equation (\ref{q(c|x)}) is independent of $c$, and the second one is non-negative. By Bayes rule, we have:
\begin{equation}
q(c | \mathbf{x}_0) = p(c|\mathbf{z}) = \frac{p(c)p(\mathbf{z}|c)}{\sum_c p(c)p(\mathbf{z}|c)},
\label{q_c_x}
\end{equation}
where $p(c)$ and $p(\mathbf{z}|c)$ have already been defined in Equations (\ref{p_c}) and (\ref{p_z_c}), respectively. This equation means that the sample $\mathbf{x}_0$ and its latent code $\mathbf{z}$ have consistent cluster assignments, which is intuitive.

So far, we have parameterized the generative model (defined in Equation (\ref{p_joint})) and the approximated posterior (defined in Equation (\ref{q_joint})) using the parameters $\left\{\boldsymbol{\pi},\, \boldsymbol{\mu}_c,\, \boldsymbol{\sigma}_c,\, \boldsymbol{\theta},\, \boldsymbol{\phi}\right\}$, $c\in \left\{1,\, 2,\, \cdots,\, k\right\}$. By substituting them into Equation (\ref{elbo}) again, the loss function of our ClusterDDPM can be derived as (see detailed derivation in supplementary):
\begin{alignat}{2}
&\mathcal{L}&&(\mathbf{x}_0) \nonumber \\
&= &&
\,E_{t, \boldsymbol{\epsilon}}\left[\left\|\boldsymbol{\epsilon} - \boldsymbol{\epsilon}_{\boldsymbol{\theta}}\left(\mathbf{x}_t, t, \mathbf{z}\right)\right\|^2\right] + \lambda D_{KL}\left(q(c | \mathbf{x}_0) || p(c)\right) \nonumber \\[2mm]
& &&
+ \lambda \sum_{c=1}^K w_c D_{KL}(q_{\boldsymbol{\phi}}(\mathbf{z} | \mathbf{x}_0) || p(\mathbf{z}|c)) \label{loss} \\[2mm]
&=  &&
\,E_{t, \boldsymbol{\epsilon}}\left[\left\|\boldsymbol{\epsilon} - \boldsymbol{\epsilon}_{\boldsymbol{\theta}}\left(\mathbf{x}_t, t, \mathbf{z}\right)\right\|^2\right] - \lambda\sum_{c=1}^K w_c \log \frac{\pi_c}{w_c} \nonumber\\[2mm]
& &&
+ \frac{\lambda}{2} \sum_{c=1}^K w_c \sum_{j=1}^J(\left.\log \boldsymbol{\sigma}_c^2\right|_j + \frac{\left.\boldsymbol{\sigma}_{\boldsymbol{\phi}}^2\right|_j}{\left.\boldsymbol{\sigma}_c^2\right|_j} +
\frac{(\left.\boldsymbol{\mu}_{\boldsymbol{\phi}}\right|_j-
\left.\boldsymbol{\mu}_c\right|_j)^2}{\left.\boldsymbol{\sigma}_c^2\right|_j}) \nonumber \\
& &&
- \frac{\lambda}{2} \sum_{j=1}^J\left(1+\left.\log \boldsymbol{\sigma}_{\boldsymbol{\phi}}^2\right|_j\right),
\label{loss_2}
\end{alignat}
where $\lambda$ is a tradeoff hyperparameter to balance the noise reconstruction loss and the prior matching loss, $J$ is the dimensionality of $\mathbf{z}$, $\ast|_j$ denotes the $j$-th element of $\ast$, and $w_c$ denotes $q(c | \mathbf{x}_0)$ for simplicity. By minimizing $\mathcal{L}(\mathbf{x}_0)$, the first term in Equation (\ref{loss}) will encourage the encoder network $\boldsymbol{f_{\phi}}$ to encode the information of $\mathbf{x}_0$ in $\mathbf{z}$, since $\mathbf{x}_t = \sqrt{\bar{\alpha}_t} \mathbf{x}_0+\sqrt{1-\bar{\alpha}_t} \boldsymbol{\epsilon}$
and the information of $\mathbf{x}_0$ can assist $\boldsymbol{\epsilon}_{\boldsymbol{\theta}}$ in more accurately predicting the noise $\boldsymbol{\epsilon}$. The rest will regularize the latent representation $\mathbf{z}$ to reside on a mixture of Gaussians manifold.

\subsection{The learning algorithm}
To minimize the loss function $\mathcal{L}(\mathbf{x}_0)$, we tailor an EM algorithm for ClusterDDPM, as simultaneously minimizing $\mathcal{L}(\mathbf{x}_0)$ w.r.t. all parameters is challenging. Specifically, we update the parameters $\left\{\boldsymbol{\pi},\, \boldsymbol{\mu}_c,\, \boldsymbol{\sigma}_c\right\}$ and $\left\{\boldsymbol{\theta},\, \boldsymbol{\phi}\right\}$ alternately to minimize $\mathcal{L}(\mathbf{x}_0)$.

\subsubsection{E-step.}
In this step, we first use the trained encoder to extract latent representations for all real samples, and then update the parameters $\left\{\boldsymbol{\pi},\, \boldsymbol{\mu}_c,\, \boldsymbol{\sigma}_c\right\}$ by grouping them with the Gaussian mixture model (GMM). Finally, the updated $\left\{\boldsymbol{\pi},\, \boldsymbol{\mu}_c,\, \boldsymbol{\sigma}_c\right\}$ are passed to the M-step for guiding the training of the networks in the next training round.

\subsubsection{M-step.}
Given the mixture of Gaussian priors $\left\{\boldsymbol{\pi},\, \boldsymbol{\mu}_c,\, \boldsymbol{\sigma}_c\right\}$ obtained from the E-step, we update $\left\{\boldsymbol{\theta},\, \boldsymbol{\phi}\right\}$ using SGD to minimize $\mathcal{L}(\mathbf{x}_0)$. Then, the trained encoder is passed to the E-step to extract latent representations.

So far, the whole training loop of ClusterDDPM has been built up. The final clustering result can be obtained by performing GMM on the learned latent representations. Note that the cluster assignment formula of GMM in the latent space is exactly Equation (\ref{q_c_x}).

\subsection{The theoretical analysis}
For a comprehensive understanding of the proposed EM framework, we provide a theoretical analysis of the optimizations performed in the M-step. We prove that maximizing $\mathcal{L}_{\mathrm{ELBO}}(\mathbf{x}_0)$ (defined in Equation (\ref{elbo})) is equivalent to maximizing the lower bound of the Q function \cite{bishop2006pattern} when $q(c | \mathbf{x}_0)$ is a fixed constant received from the E-step. The proof is as follows:
\begin{alignat}{2}
&\mathcal{L}&&_{\mathrm{ELBO}}(\mathbf{x}_0)
\nonumber \\
&= &&\,
E_{q_{\boldsymbol{\phi}}(\mathbf{x}_{1:T}, \mathbf{z}, c | \mathbf{x}_0)}\left[\log \frac{p\left(\mathbf{x}_{0:T}, \mathbf{z}, c\right)}{q_{\boldsymbol{\phi}}(\mathbf{x}_{1:T}, \mathbf{z}, c | \mathbf{x}_0)}\right]
\nonumber \\[2mm]
&= &&\,
E_{q_{\boldsymbol{\phi}}(\mathbf{x}_{1:T}, \mathbf{z}, c | \mathbf{x}_0)}\left[\log \frac{p\left(\mathbf{x}_{0:T}, \mathbf{z}, c\right)}{p(\mathbf{x}_{1:T}, \mathbf{z}| \mathbf{x}_0, c)} \frac{p(\mathbf{x}_{1:T}, \mathbf{z}| \mathbf{x}_0, c)}{q_{\boldsymbol{\phi}}(\mathbf{x}_{1:T}, \mathbf{z}, c | \mathbf{x}_0)}\right]
\nonumber \\[2mm]
&= &&\,
E_{q_{\boldsymbol{\phi}}(\mathbf{x}_{1:T}, \mathbf{z}, c | \mathbf{x}_0)}\left[\log p\left(\mathbf{x}_0, c\right)\right]
\nonumber \\[2mm]
& &&
- D_{KL}(q_{\boldsymbol{\phi}}(\mathbf{x}_{1:T}, \mathbf{z}, c | \mathbf{x}_0) || p(\mathbf{x}_{1:T}, \mathbf{z}| \mathbf{x}_0, c)) \nonumber \\[2mm]
&\leq &&\,
E_{q(c | \mathbf{x}_0)}\left[\log p\left(\mathbf{x}_0, c\right)\right]
\nonumber \\[2mm]
&= &&\,
\sum_{c=1}^K\left[q(c | \mathbf{x}_0)\log p\left(\mathbf{x}_0, c\right)\right].
\label{q_function}
\end{alignat}
Upon setting $q(c | \mathbf{x}_0)$ as a fixed constant received from the E-step, Equation (\ref{q_function}) becomes the Q function.

While this constraint is easily satisfied, the fixed nature of $q(c | \mathbf{x}_0)$ may result in the prior matching term $D_{KL}\left(q(c | \mathbf{x}_0) || p(c)\right)$ in Equation (\ref{loss}) becoming a constant. As a result, valuable prior probability information for clusters may be squandered. To address this, in our proposed framework, only $\left\{\boldsymbol{\pi},\, \boldsymbol{\mu}_c,\, \boldsymbol{\sigma}_c\right\}$ is passed from the E-step to the M-step, whereas $q(c | \mathbf{x}_0)$ varies with each iteration step. By allowing $q(c | \mathbf{x}_0)$ to change iteratively, the EM framework can better capture the dynamic nature of the data clustering process, incorporating prior information about cluster probabilities more effectively. 

\begin{table}[!t]
\caption{Description of datasets.}
\renewcommand{\arraystretch}{1} 
\tabcolsep 8mm 
\begin{tabular}{lccc}
\hline\hline
\textbf{Dataset} &\textbf{Size} & \textbf{Image size} & \textbf{Class} \\\hline
MNIST             & 70000    & $28\times28$   & 10\\
Fashion-MNIST     & 70000    & $28\times28$   & 10\\
CIFAR-10          & 60000    & $32\times32$   & 10 \\
COIL20            & 1440     & $128\times128$ & 20\\

\hline\hline
\label{datasets}
\end{tabular}
\end{table}

\section{Experiments}
In this section, we describe the experimental settings, validate the effectiveness of ClusterDDPM, analyze the learned latent representations qualitatively and quantitatively, and examine the sensitivity of DDPM to the tradeoff hyperparameter $\lambda$.

\subsection{Experimental settings}
To evaluate the performance of ClusterDDPM, we conduct experiments on four benchmark datasets (Table \ref{datasets}): MNIST \cite{lecun1998gradient}, Fashion-MNIST \cite{xiao2017fashion}, CIFAR-10 \cite{krizhevsky2009learning} and COIL20 \cite{nene1996columbia}.

In ClusterDDPM, the noise predictor $\boldsymbol{\epsilon}_{\boldsymbol{\theta}}$ is implemented as a conditional U-Net with adaptive group normalization \cite{dhariwal2021diffusion}, integrating timestep $t$ and latent representation $z$ into each residual block following a group normalization operation \cite{wu2018group}. The encoder $\boldsymbol{f_{\phi}}$ shares the same architecture as the first half of $\boldsymbol{\epsilon}_{\boldsymbol{\theta}}$. We train the models using the Adam optimizer \cite{kingma2014adam}. The latent representation dimensionality is set to 32, 64, 512, and 10 for MNIST, Fashion-MNIST, CIFAR-10, and COIL20, respectively. The tradeoff hyperparameter $\lambda$ is tuned to 0.1, 0.001, 0.01, and 0.05 for the respective datasets. More detailed hyperparameter settings are provided in the supplementary materials. The code will be made publicly available.


\begin{table*}[!t]
\centering
\caption{The performance of different clustering methods.}
\renewcommand{\arraystretch}{1.2} 
\tabcolsep 1.1mm 
\begin{tabular}{lcccccccc}
\hline\hline
\multirow{2}{*}{Method}
                        &\multicolumn{2}{c}{MNIST} &\multicolumn{2}{c}{Fashion-MNIST}
                        &\multicolumn{2}{c}{CIFAR-10}  &\multicolumn{2}{c}{COIL20}\\
                        \cmidrule(r){2-3} \cmidrule(r){4-5}
                        \cmidrule(r){6-7} \cmidrule(r){8-9}
\quad
                        &ACC &NMI        &ACC &NMI
                        &ACC &NMI        &ACC &NMI\\\hline

\hline
K-means & 53.24 & 0.499 & 51.08 & 0.392 & 23.6 & 0.102 & 47.33 & 0.48 \\

GMM &53.72 & 0.501 & 52.34 & 0.409 & 21.47 & 0.113 & 48.16 & 0.477\\

VaDE & 94.46 & 0.891 & 62.87 & 0.611 & 26.65 & 0.151 & 75.48 & 0.779\\

ClusterGAN &95 & 0.89 & 63 & 0.64 & 28.12 & 0.156 & 77 & 0.811\\
GAN-EM &94.93 & 0.899 & 63.57 & 0.642 & 23.14 & 0.11 & 74.03 & 0.833\\
HC-MGAN &94.3 & 0.905 & \textbf{72.1} & \textbf{0.691} & - & - & - & - \\
VaGAN-GMM &95.48 & 0.917 & 63.84 & 0.633 & 28.79 & 0.158 & 79.2 & 0.851\\
VaGAN-SMM &95.75 & 0.901 & 66.95 & 0.671 & 30.2 & 0.172 & 81.67 & 0.868 \\
ClusterDPMM &\textbf{97.67} & \textbf{0.94} & 70.49 & 0.661 & \textbf{30.46} & \textbf{0.173} & \textbf{82.43} & \textbf{0.906}\\
\hline\hline
\end{tabular}\label{performance}
\end{table*}

\subsection{Effectiveness analysis of ClusterDDPM}
To showcase the advantages of ClusterDDPM, we compare it against eight baseline models, including two classic clustering methods: K-means and GMM, and six state-of-the-art deep generative models: variational deep embedding (VaDE) \cite{vade}, ClusterGAN \cite{mukherjee2019clustergan}, GAN-based EM learning framework (GAN-EM) \cite{gan-em}, hierarchical clustering with multi-generator GANs (HC-MGAN) \cite{de2022top}, VaGAN-GMM \cite{yang2020clustering} and VaGAN-SMM \cite{yang2020clustering}.

\begin{figure}[!t]
\centering
\subfigure[GMM]{
\includegraphics[height = 4.cm, width = 4.5cm]{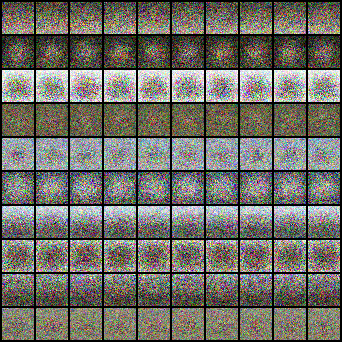}}
\quad
\subfigure[VaDE]{
\includegraphics[height = 4.cm, width = 4.5cm]{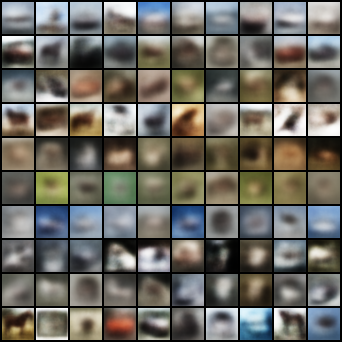}}

\subfigure[VaGAN-GMM]{
\includegraphics[height = 4.cm, width = 4.5cm]{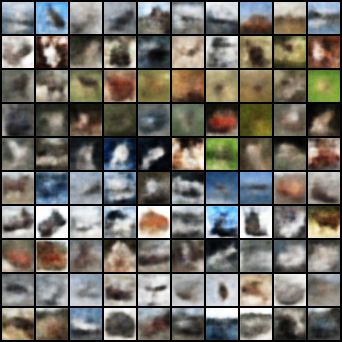}}
\quad
\subfigure[Ours]{
\includegraphics[height = 4.cm, width = 4.5cm]{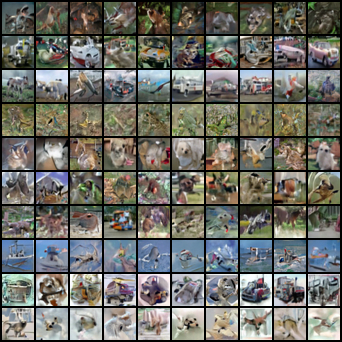}}

\caption{Unsupervised conditional generation on CIFAR-10. Each row corresponds to a cluster.}
\label{vis_fake}
\end{figure}

As shown in Table \ref{performance}, one can see that: 1) ClusterDDPM outperforms the baseline models significantly on most datasets. 2) Improved generative performance and representation learning abilities positively impact clustering performance. For instance, ClusterDDPM outperforms VaGAN-GMM, and VaGAN-GMM in turn outperforms VaDE. While VaDE and VaGAN-GMM are the most relevant methods to ClusterDDPM, their generative models' effectiveness falls behind the DDPM, limiting their potential in clustering performance and unsupervised conditional generation. As shown in Fig. \ref{vis_fake}, samples generated by the proposed model exhibit superior clarity, and the intra-cluster generated samples showcase a more conspicuous shared trait, i.e. akin backgrounds. 



\begin{figure*}[!t]
\centering
\subfigure[Original data space]{
\includegraphics[height = 4.15cm, width = 4.5cm]{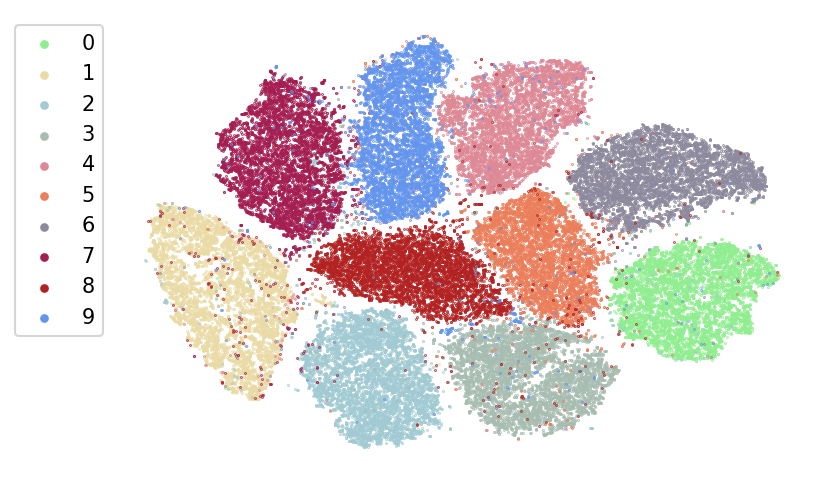}}
\quad

\subfigure[VaDE]{
\includegraphics[height = 4.15cm, width = 4.cm]{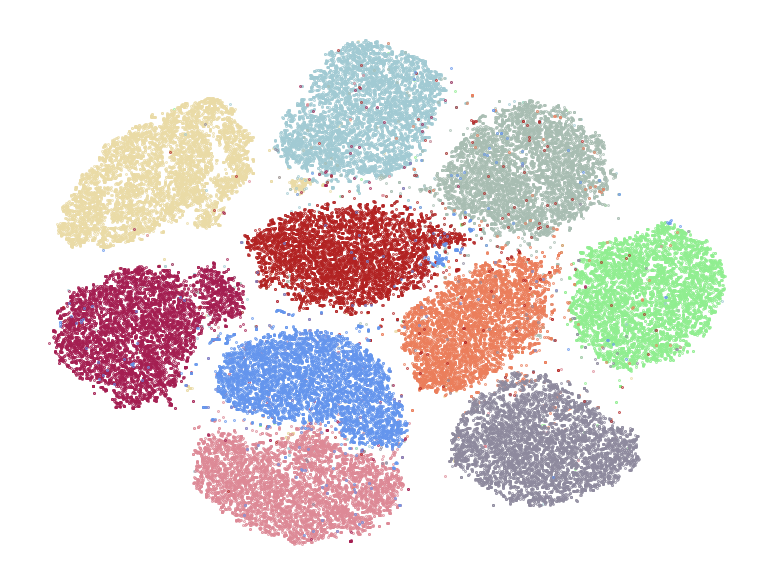}}
\quad
\subfigure[VaGAN-GMM]{
\includegraphics[height = 4.15cm, width = 4.cm]{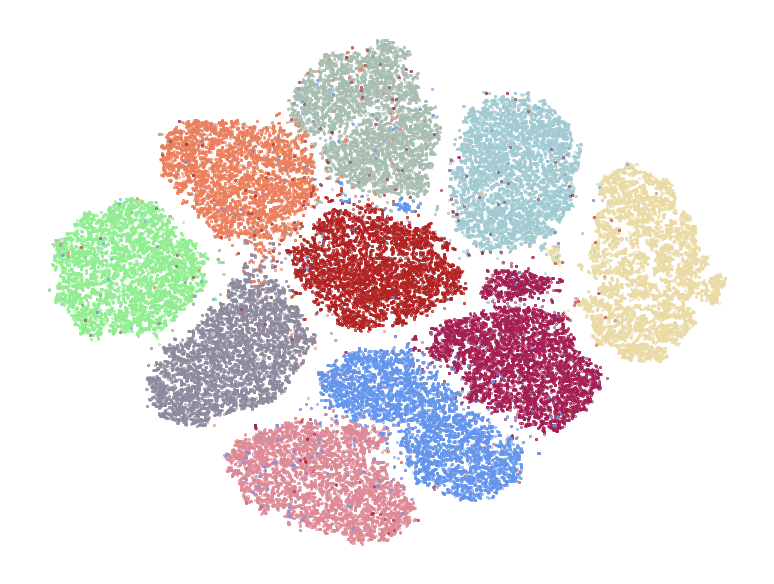}}
\quad
\subfigure[Ours]{
\includegraphics[height = 4.15cm, width = 4.cm]{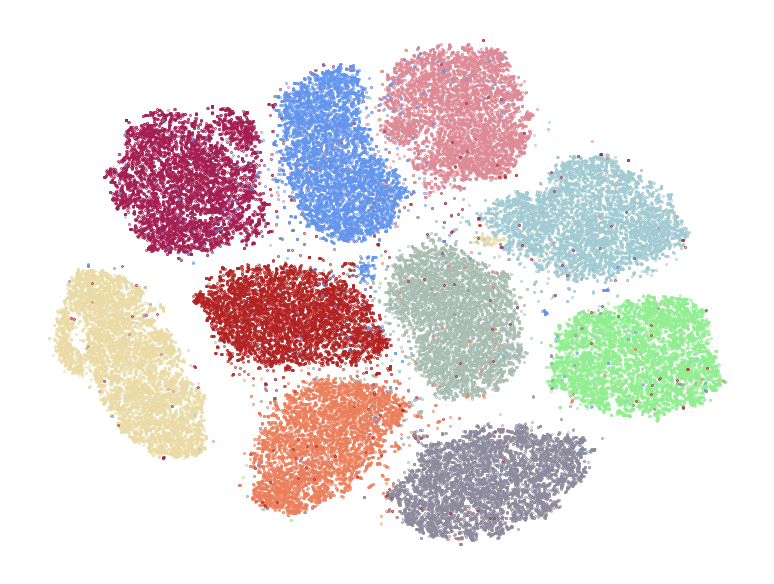}}

\subfigure[VaDE]{
\includegraphics[height = 4.15cm, width = 4.cm]{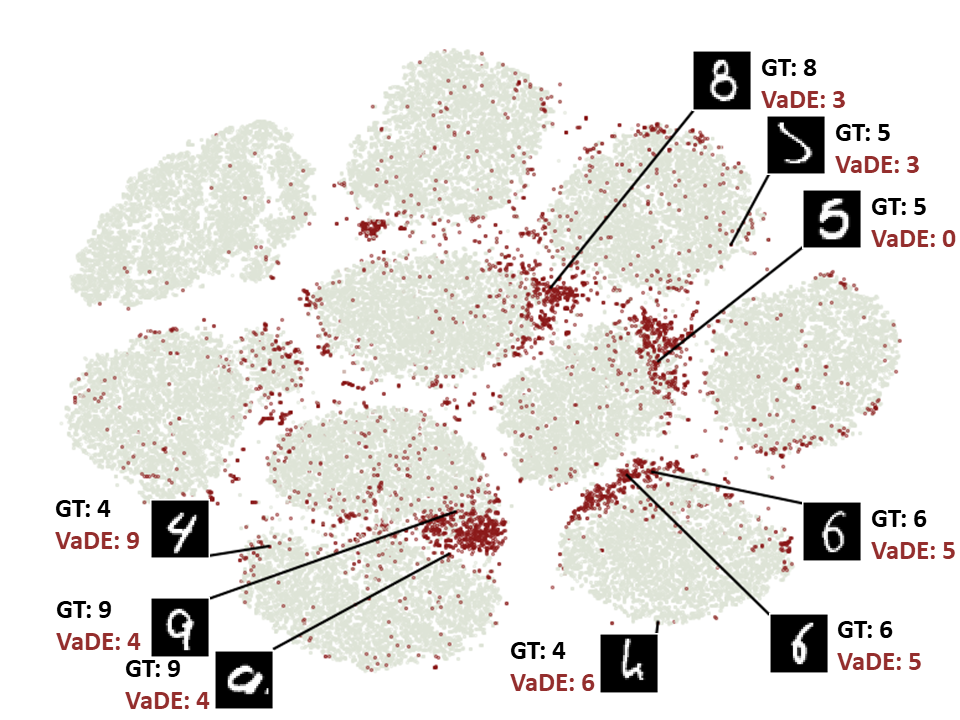}}
\quad
\subfigure[VaGAN-GMM]{
\includegraphics[height = 4.15cm, width = 4.cm]{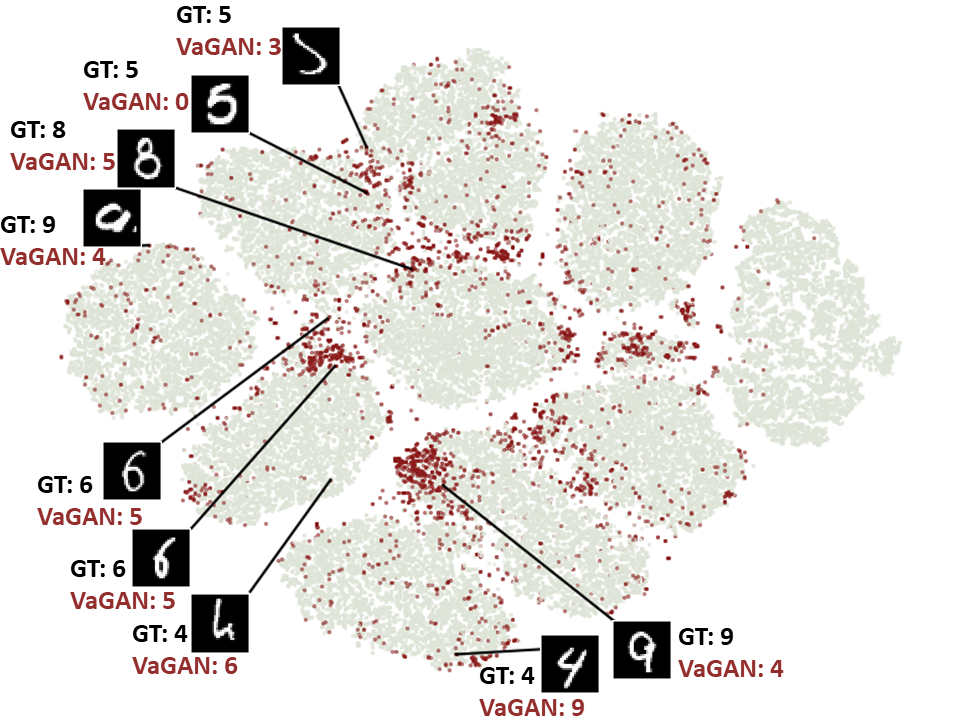}}
\quad
\subfigure[Ours]{
\includegraphics[height = 4.15cm, width = 4.cm]{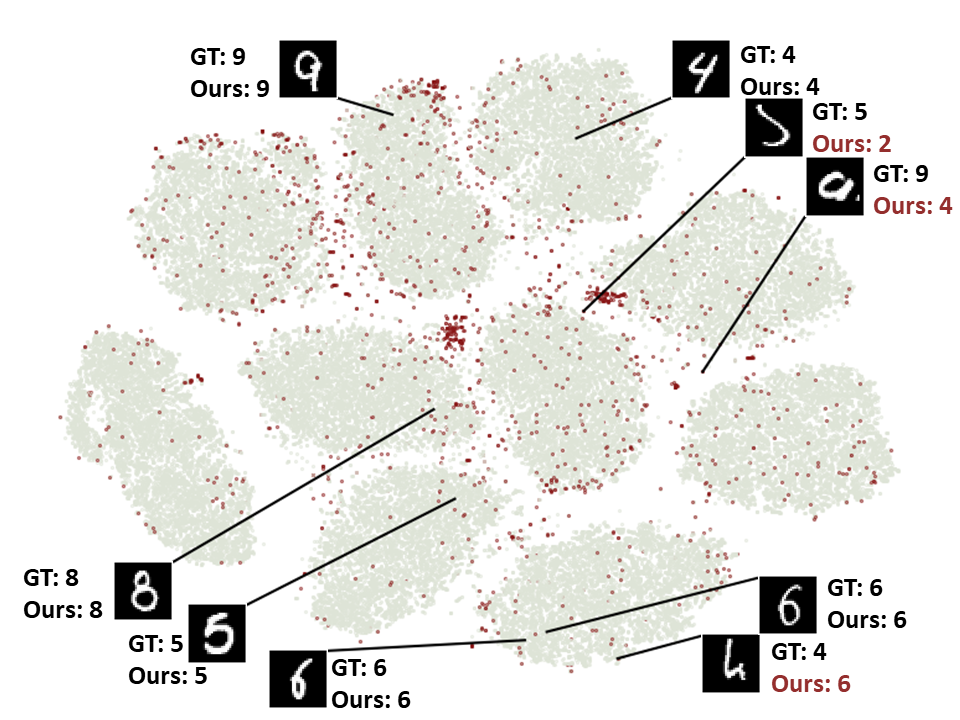}}

\caption{t-SNE visualization on the dataset MNIST (best viewed in color). (a) shows the samples in the original data space, where each color corresponds to a specific category of digits and many samples from different categories are mixed together. (b)-(d) show the samples in different latent spaces, where the mixing problem is noticeably alleviated. (e)-(f) show the misclustered samples in different latent spaces, marked in red. "GT:4" signifies the ground-truth label of the adjacent image is 4. "VaDE: 4" denotes the predicted one obtained by VaDE is 4, and so on.}
\label{case1}
\end{figure*}

\subsection{Effectiveness analysis of the learned latent representations}
In this section, we analyze the superior performance of ClusterDDPM by qualitatively and quantitatively analyzing the effectiveness of the learned latent representations.

\subsubsection{Qualitative analysis.}
For qualitative analysis, we first use t-SNE \cite{van2008visualizing} to visualize the learned latent representations of VaDE, VaGAN-GMM, and ClusterDDPM on the MNIST dataset. Then, we show the misclustered samples in different latent spaces.

As shown in Fig. \ref{case1}, it reveals the following observations: 1) In the original data space, clear modes are evident, but many samples from different categories are mixed, making clustering challenging. 2) Latent representation learning significantly alleviates the mixing problem, and the latent representations of the proposed model are more GMM-friendly compared to the baselines. 3) Most misclustered samples are located near at the border of each category. 4) Notably, the proposed model exhibits a significantly lower number of misclustered points near category boundaries compared to the baselines. The samples situated at category borders often lack typical category characteristics; otherwise, they would be surrounded by many similar points. However, some clear digits with typical category characteristics are mapped to the category boundaries by the baselines, resulting in misclustering. In contrast, the proposed model maps them to the interior of the categories and recovers their ground-truth labels, indicating that it better preserves the inherent structure of the data.

\subsubsection{Quantitative analysis.}
In the quantitative analysis, we run a k-Nearest Neighbors classifier (kNN) on MNIST to assess the discriminative power of the learned latent representations, following \cite{nalisnick2017stickbreaking, vade}. A better test performance indicates fewer samples of different categories among the neighbors and less severe mixing problem. 

Table \ref{case 2} reveals the following findings: 1) The classification performance on the latent space is comparable to or worse than that on the original data space when k is small (k = 3,\, 5,\, 7,\, 9). However, as k increases to 100, 1000, and 6000, the advantages of latent representation learning become evident, effectively alleviating the mixing problem in the original data space. 2) The learned latent representations of the proposed model are more effective and robust compared to the baselines, preserving the inherent structure of the data better.

\begin{table}[!t]
\caption{MNIST test accuracy (\%) for kNN on different latent spaces.}
\renewcommand{\arraystretch}{1.2} 
\tabcolsep 2.2mm 
\begin{tabular}{lccccccc}
\hline\hline
k  &3 &5 &7 &9 &100 &1000 &6000  \\\hline
Raw pixels & 97.05 & 96.88 & 96.94 & 96.59 & 94.4 & 87.31 & 72.13                        \\
VaDE & 97.51 & 97.69 & 97.69 & 97.7 & 97.27 & 95.77 & 92.5                             \\
VaGAN-GMM & 96.81 & 97.02 & 97.07 & 97.21 & 96.88 & 95.84 & 93.32                       \\
Ours & \textbf{97.62} & \textbf{97.99} & \textbf{98.02} & \textbf{98.03} & \textbf{97.93} & \textbf{97.33} & \textbf{96.74}                             \\
\hline\hline
\end{tabular}
\label{case 2}
\end{table}

\begin{table}[!t]
\caption{The correlation between the hpyerparameter $\lambda$ and the clustering accuracy (\%) of ClusterDDPM. Blank entries indicate that the corresponding model trainings are unstable.}
\renewcommand{\arraystretch}{1.2} 
\tabcolsep 3.2mm 
\begin{tabular}{lcccccc}
\hline\hline
$\lambda$  & 0.0001 & 0.001 & 0.01 & 0.05 & 0.1 & 0.5  \\\hline
MNIST & 67.69 & 69.38 & 82.52 & 91.97 & \textbf{95.93} & 30.73 \\
Fashion-MNIST & 67.49 & \textbf{70.49} & 68.14 & 59.44 & - & - \\
CIFAR-10 & 27.79 & 27.58 & \textbf{30.46} & - & - & - \\
COIL20 & 80.21 & 80.9 & 81.18 & \textbf{82.43} & 81.39 & 79.31 \\
\hline\hline
\end{tabular}
\label{case 3}
\end{table}

\subsection{Hyperparameter sensitivity}

In this section, we analyze the sensitivity of clustering performance to the tradeoff hyperparameter $\lambda$ on different datasets. For each comparsion, apart from $\lambda$, we keep all other experimental settings the same for ClusterDDPM.

\begin{figure}[!t]
\centering
\subfigure{
\includegraphics[height = 4cm, width = 6.25cm]{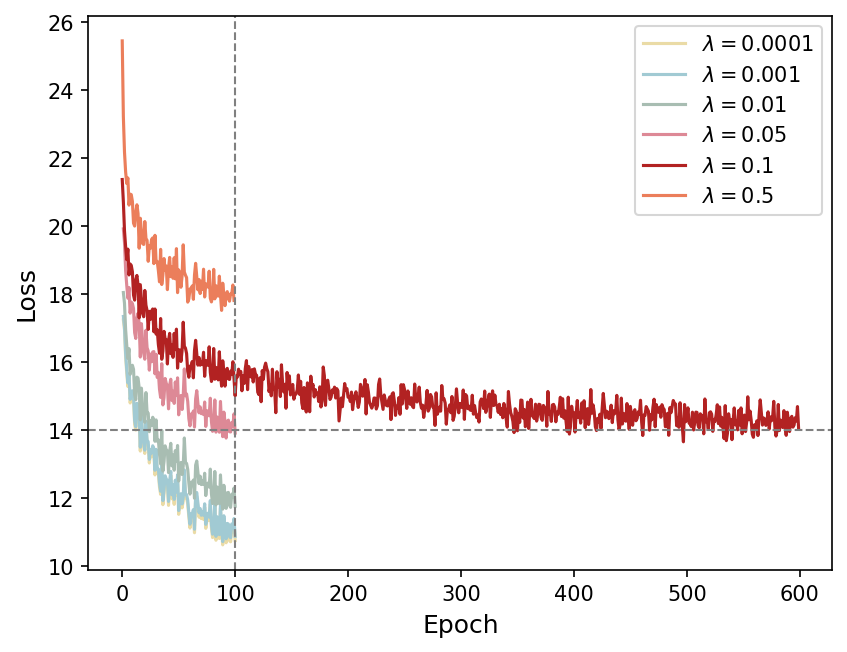}}
\quad
\subfigure{
\includegraphics[height = 4cm, width = 6.25cm]{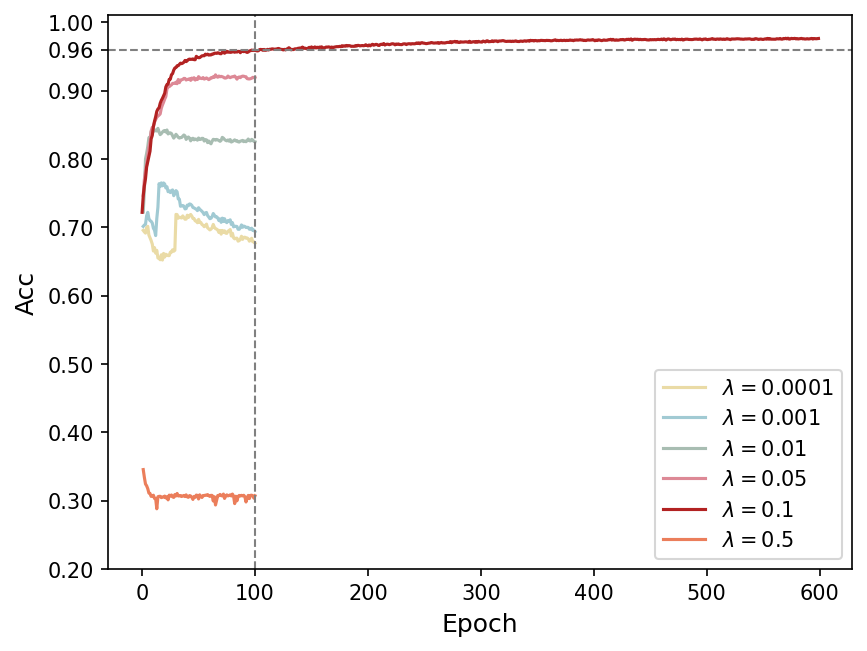}}

\caption{Convergence of ClusterDDPM with different $\lambda$ on MNIST.}
\label{case3_loss}
\end{figure}

Table \ref{case 3} and Figure \ref{case3_loss} illustrate the following observations: 1) Prior matching terms in Equation (\ref{loss}) contribute to improved clustering. A smaller $\lambda$ causes the prior matching terms to have a minor influence on the gradient direction during training, leading to faster convergence and smaller model loss values, but potentially unsatisfactory clustering results. A slight increase in $\lambda$ significantly improves clustering performance. 2) Avoiding excessively large $\lambda$ is crucial, as prior matching terms may dominate the gradient direction, leading to latent representations that fail to preserve the inherent data structure and result in poor clustering performance and unstable training. Overall, the recommended range for $\lambda$ is [0.001, 0.1].

\section{Conclusion}
In this study, we proposed ClusterDDPM, an innovative EM clustering framework with DDPMs. As far as we can ascertain, this is the first exploration of DDPMs in clustering. Comprehensive experiments validate the advantages of the proposed model in terms of clustering, unsupervised conditional generation and latent representation learning.  Furthermore, we provided a theoretical analysis of the optimization process in the M-step, and proved that the optimizations are equivalent to maximizing the lower bound of the Q function within the vanilla EM framework under certain constraints.

While our experiments are conducted with the most basic architecture of DDPMs, our ClusterDDPM framework lays a strong foundation for incorporating more effective variations of DDPMs, which holds the promise of achieving even better results in future research.

\appendix
\section{Derivation of the objective function}
By substituting the defined generative model $p\left(\mathbf{x}_{0:T}, \mathbf{z}, c\right)$ and the defined approximated posterior $q_{\boldsymbol{\phi}}(\mathbf{x}_{1:T}, \mathbf{z}, c | \mathbf{x}_0)$ into the evidence lower bound (ELBO) of the log-likelihood of ClusterDDPM, the ELBO can be reformulated as:
\begin{alignat}{2}
&\mathcal{L}&&_{\mathrm{ELBO}}(\mathbf{x}_0)
\nonumber \\
&= &&\,
E_{q_{\boldsymbol{\phi}}(\mathbf{x}_{1:T}, \mathbf{z}, c | \mathbf{x}_0)}\left[\log \frac{p\left(\mathbf{x}_{0:T}, \mathbf{z}, c\right)}{q_{\boldsymbol{\phi}}(\mathbf{x}_{1:T}, \mathbf{z}, c | \mathbf{x}_0)}\right] \\[2mm]
&= &&\,
E_{q_{\boldsymbol{\phi}}(\mathbf{x}_{1:T}, \mathbf{z}, c | \mathbf{x}_0)}\left[\log \frac{p\left(\mathbf{x}_T\right) \prod_{t=1}^T p_{\boldsymbol{\theta}}\left(\mathbf{x}_{t-1} | \mathbf{x}_t, \mathbf{z}\right)}{q(\mathbf{x}_{1:T} | \mathbf{x}_0)} \right] \nonumber\\[2mm]
& &&+
E_{q_{\boldsymbol{\phi}}(\mathbf{x}_{1:T}, \mathbf{z}, c | \mathbf{x}_0)}\left[\log \frac{p(\mathbf{z}, c)}{q_{\boldsymbol{\phi}}(\mathbf{z}, c | \mathbf{x}_0)} \right]\\[2mm]
&= &&\,
E_{q_{\boldsymbol{\phi}}(\mathbf{x}_{1:T},\, \mathbf{z} | \mathbf{x}_0)}\left[\log \frac{p\left(\mathbf{x}_T\right) \prod_{t=1}^T p_{\boldsymbol{\theta}}\left(\mathbf{x}_{t-1} | \mathbf{x}_t, \mathbf{z}\right)}{\prod_{t=1}^T q\left(\mathbf{x}_{t} | \mathbf{x}_{t - 1}\right)} \right] \nonumber\\[2mm]
& &&-
D_{KL}(q_{\boldsymbol{\phi}}(\mathbf{z}, c | \mathbf{x}_0) || p(\mathbf{z}, c)).
\end{alignat}
Similar to the DDPM \cite{ho2020denoising, luo2022understanding}, the reconstruction term can be further derived as:
\begin{align}
&E_{q_{\boldsymbol{\phi}}(\mathbf{x}_{1:T},\, \mathbf{z} | \mathbf{x}_0)}\left[\log \frac{p\left(\mathbf{x}_T\right) \prod_{t=1}^T p_{\boldsymbol{\theta}}\left(\mathbf{x}_{t-1} | \mathbf{x}_t, \mathbf{z}\right)}{\prod_{t=1}^T q\left(\mathbf{x}_{t} | \mathbf{x}_{t - 1}\right)} \right] \nonumber \\[2mm]
&=L_0 - L_T - \sum_{t=2}^T L_{t-1}, \\[2mm]
L_0 & = E_{q_{\boldsymbol{\phi}}(\mathbf{x}_{1},\, \mathbf{z} | \mathbf{x}_0)}\left[\log p_{\boldsymbol{\theta}}\left(\mathbf{x}_{0} | \mathbf{x}_1, \mathbf{z}\right)\right], \\[2mm]
L_T &= D_{KL}(q(\mathbf{x}_{T} | \mathbf{x}_0) || p(\mathbf{x}_{T})), \\[2mm]
L_{t-1} &= E_{q_{\boldsymbol{\phi}}(\mathbf{x}_{t}, \mathbf{z} | \mathbf{x}_0)}\left[ D_{KL}(q(\mathbf{x}_{t-1} | \mathbf{x}_{t}, \mathbf{x}_{0}) || p_{\boldsymbol{\theta}}\left(\mathbf{x}_{t-1} | \mathbf{x}_t, \mathbf{z}\right) \right].
\end{align}
The $L_T$ can be disregarded, as it does not possess any trainable parameters. The $L_{t - 1}$ can be evaluated in closed form. With $q\left(\mathbf{x}_{t-1} | \mathbf{x}_{t}, \mathbf{x}_0\right) = \mathcal{N}\left(\mathbf{x}_{t-1} | \boldsymbol{\mu}_q\left(\mathbf{x}_t, \mathbf{x}_0\right), \boldsymbol{\Sigma}_q\left(t\right)\right)$ and $p_{\boldsymbol{\theta}}\left(\mathbf{x}_{t-1} | \mathbf{x}_t, \mathbf{z}\right) = \mathcal{N}\left(\mathbf{x}_{t-1} | \boldsymbol{\mu}_{\boldsymbol{\theta}}\left(\mathbf{x}_t, t, \mathbf{z}\right), \beta_t\mathbf{I}\right)$, $L_{t - 1}$ can be derived as:
\begin{align}
L_{t-1} &= E_{q_{\boldsymbol{\phi}}(\mathbf{x}_{t}, \mathbf{z} | \mathbf{x}_0)}\left[\frac{1}{2 \beta_t}\left\|\boldsymbol{\mu}_q\left(\mathbf{x}_t, \mathbf{x}_0\right) -\boldsymbol{\mu}_{\boldsymbol{\theta}}\left(\mathbf{x}_t, t, \mathbf{z}\right)\right\|^2\right] \nonumber\\[2mm]
& \quad + C \\[2mm]
&= E_{q_{\boldsymbol{\phi}}(\mathbf{x}_{t}, \mathbf{z} | \mathbf{x}_0)} [\frac{1}{2 \beta_t}\|\frac{1}{\sqrt{\alpha_t}}(\mathbf{x}_t-\frac{\beta_t}{\sqrt{1-\bar{\alpha}_t}} \boldsymbol{\epsilon}) \nonumber \\[2mm]
&\quad - \frac{1}{\sqrt{\alpha_t}}(\mathbf{x}_t-\frac{\beta_t}{\sqrt{1-\bar{\alpha}_t}} \boldsymbol{\epsilon}_{\boldsymbol{\theta}}(\mathbf{x}_t, t, \mathbf{z}))\|^2] + C \\[2mm]
&= E_{q_{\boldsymbol{\phi}}(\mathbf{x}_{t}, \mathbf{z} | \mathbf{x}_0)} [\frac{\beta_t}{2 \alpha_t (1-\bar{\alpha}_t)}\|\boldsymbol{\epsilon} - \boldsymbol{\epsilon}_{\boldsymbol{\theta}}(\mathbf{x}_t, t, \mathbf{z}))\|^2] + C, \label{denoising_match}
\end{align}
where $\epsilon$ is the noise added to $\mathbf{x}_0$ to obtain $\mathbf{x}_t$ and $\boldsymbol{\epsilon} \sim \mathcal{N}\left(\mathbf{0}, \, \mathbf{I}\right)$, and $C$ is a constant. Following the DDPM \cite{ho2020denoising}, we simplify Equation (\ref{denoising_match}) into an unweighted version:
\begin{align}
L_{t-1} &= E_{q_{\boldsymbol{\phi}}(\mathbf{x}_{t}, \mathbf{z} | \mathbf{x}_0)} [\|\boldsymbol{\epsilon} - \boldsymbol{\epsilon}_{\boldsymbol{\theta}}(\mathbf{x}_t, t, \mathbf{z}))\|^2].
\end{align}
And the $t = 1$ case corresponds to the negative $L_{0}$. Thus, the reconstruction term can be rewritten as:
\begin{align}
&E_{q_{\boldsymbol{\phi}}(\mathbf{x}_{1:T},\, \mathbf{z} | \mathbf{x}_0)}\left[\log \frac{p\left(\mathbf{x}_T\right) \prod_{t=1}^T p_{\boldsymbol{\theta}}\left(\mathbf{x}_{t-1} | \mathbf{x}_t, \mathbf{z}\right)}{\prod_{t=1}^T q\left(\mathbf{x}_{t} | \mathbf{x}_{t - 1}\right)} \right] \nonumber \\[2mm]
&\propto - \sum_{t=1}^T E_{q_{\boldsymbol{\phi}}(\mathbf{x}_{t}, \mathbf{z} | \mathbf{x}_0)} \left[\|\boldsymbol{\epsilon} - \boldsymbol{\epsilon}_{\boldsymbol{\theta}}(\mathbf{x}_t, t, \mathbf{z}))\|^2\right]. \label{e12}
\end{align}
For the prior matching term, it can be derived as \cite{vade}:
\begin{align}
& D_{KL}(q_{\boldsymbol{\phi}}(\mathbf{z}, c | \mathbf{x}_0) || p(\mathbf{z}, c))\nonumber \\[2mm]
&=
D_{KL}\left(q(c | \mathbf{x}_0) || p(c)\right) + \sum_{c=1}^K w_c D_{KL}(q_{\boldsymbol{\phi}}(\mathbf{z} | \mathbf{x}_0) || p(\mathbf{z}|c)) \\[2mm]
&= - \sum_{c=1}^K w_c \log \frac{\pi_c}{w_c} - \frac{1}{2} \sum_{j=1}^J\left(1+\left.\log \boldsymbol{\sigma}_{\boldsymbol{\phi}}^2\right|_j\right) \\[2mm]
& \quad + \frac{1}{2} \sum_{c=1}^K w_c \sum_{j=1}^J(\left.\log \boldsymbol{\sigma}_c^2\right|_j + \frac{\left.\boldsymbol{\sigma}_{\boldsymbol{\phi}}^2\right|_j}{\left.\boldsymbol{\sigma}_c^2\right|_j} + \frac{(\left.\boldsymbol{\mu}_{\boldsymbol{\phi}}\right|_j-
\left.\boldsymbol{\mu}_c\right|_j)^2}{\left.\boldsymbol{\sigma}_c^2\right|_j}).
\end{align}

For one thing, Equation (\ref{e12}) is an approximation of the construction term. For another, it is difficult to calculate the noise reconstruction loss for each sample over all time steps in a training batch, as $T$ is often quite large (e.g., 1000). To handle this, we only calculate the noise reconstruction loss for each sample over a single time step in a training batch, and use the tradeoff hyperparameter $\lambda$ to balance the noise reconstruction loss and the prior matching loss. Thus the loss function of our ClusterDDPM can be defined as:
\begin{alignat}{2}
&\mathcal{L}&&(\mathbf{x}_0) \nonumber \\[2mm]
&= && - \mathcal{L}_{\mathrm{ELBO}}(\mathbf{x}_0)\nonumber \\[2mm]
&= &&
\,E_{t, \boldsymbol{\epsilon}}\left[\left\|\boldsymbol{\epsilon} - \boldsymbol{\epsilon}_{\boldsymbol{\theta}}\left(\mathbf{x}_t, t, \mathbf{z}\right)\right\|^2\right] + \lambda D_{KL}\left(q(c | \mathbf{x}_0) || p(c)\right) \nonumber \\[2mm]
& &&
+ \lambda \sum_{c=1}^K w_c D_{KL}(q_{\boldsymbol{\phi}}(\mathbf{z} | \mathbf{x}_0) || p(\mathbf{z}|c)) \label{loss} \\[2mm]
&=  &&
\,E_{t, \boldsymbol{\epsilon}}\left[\left\|\boldsymbol{\epsilon} - \boldsymbol{\epsilon}_{\boldsymbol{\theta}}\left(\mathbf{x}_t, t, \mathbf{z}\right)\right\|^2\right] - \lambda\sum_{c=1}^K w_c \log \frac{\pi_c}{w_c} \nonumber\\[2mm]
& &&
+ \frac{\lambda}{2} \sum_{c=1}^K w_c \sum_{j=1}^J(\left.\log \boldsymbol{\sigma}_c^2\right|_j + \frac{\left.\boldsymbol{\sigma}_{\boldsymbol{\phi}}^2\right|_j}{\left.\boldsymbol{\sigma}_c^2\right|_j} +
\frac{(\left.\boldsymbol{\mu}_{\boldsymbol{\phi}}\right|_j-
\left.\boldsymbol{\mu}_c\right|_j)^2}{\left.\boldsymbol{\sigma}_c^2\right|_j}) \nonumber \\
& &&
- \frac{\lambda}{2} \sum_{j=1}^J\left(1+\left.\log \boldsymbol{\sigma}_{\boldsymbol{\phi}}^2\right|_j\right).
\end{alignat}

\section{Experimental details}
In ClusterDDPM, the noise predictor $\boldsymbol{\epsilon}_{\boldsymbol{\theta}}$ is implemented as a conditional U-Net with adaptive group normalization \cite{dhariwal2021diffusion}, integrating timestep $t$ and latent representation $z$ into each residual block following a group normalization operation \cite{wu2018group}. The encoder $\boldsymbol{f_{\phi}}$ shares the same architecture as the first half of $\boldsymbol{\epsilon}_{\boldsymbol{\theta}}$. Specifically, our MNIST model has 2 million parameters, our Fashion-MNIST model has 2.01 million parameters, our COIL20 model has 11.14 million parameters and our CIFAR-10 model has 46.2 million parameters. We train the CIFAR-10 model using 2 A5000 GPUs, while the remaining models are trained using a single 3080 GPU.

\section{Additional generated samples}

\begin{figure*}[!t]
\centering
\subfigure[MNIST]{
\includegraphics[height = 4.cm, width = 4.cm]{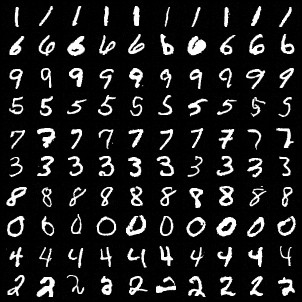}}
\quad
\subfigure[Fashion-MNIST]{
\includegraphics[height = 4cm, width = 4.cm]{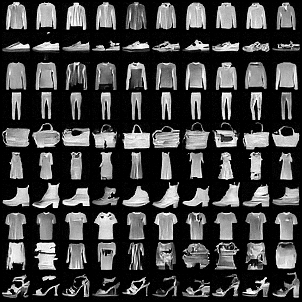}}
\quad
\subfigure[COIL20]{
\includegraphics[height = 4.cm, width = 4.cm]{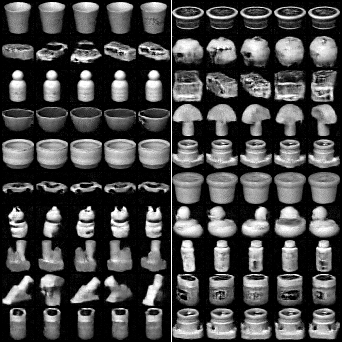}}
\caption{Unsupervised conditional generation on MNIST, Fashion-MNIST and COIL20. }
\label{cluser_gen}
\end{figure*}

\begin{figure*}[!t]
\centering
\includegraphics[height = 3cm, width = 13.5cm]{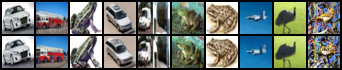}
\caption{Reconstruction results on CIFAR-10. The first row displays real images, while the second row illustrates the reconstruction results obtained by conditioning on the latent representations of the images in the first row.}
\label{rec}
\end{figure*}

Fig. \ref{cluser_gen} shows uncurated samples from our ClusterDDPM trained on MNIST, Fashion-MNIST and COIL20 datasets. Fig. \ref{rec} shows the reconstructed results from our ClusterDDPM trained on CIFAR-10 dataset.

\bibliographystyle{elsarticle-num}
\bibliography{references}
\end{document}